\def\year{2021}\relax
\documentclass[letterpaper]{article} 
\usepackage{aaai21}  
\usepackage{times}  
\usepackage{helvet} 
\usepackage{courier}  
\usepackage[hyphens]{url}  
\usepackage{graphicx} 
\usepackage{natbib}  
\usepackage{caption} 
\usepackage{subcaption}
\usepackage{float}
\usepackage{amsmath}
\usepackage{dsfont}

\usepackage{xcolor}

\usepackage{cite}
\usepackage[ruled,vlined]{algorithm2e}

\title{Detecting and Adapting to Novelty in Games}
\author{
    Xiangyu Peng,
    Jonathan C. Balloch,
    Mark O. Riedl
    \\
}

\affiliations{
    Georgia Institute of Technology \\
    \{xpeng62, balloch, riedl\}@gatech.edu

}

\begin{document}

\maketitle

\begin{abstract}
Open-world novelty occurs when the rules of an environment can change abruptly, such as when a game player encounters ``house rules''. 
To address open-world novelty, game playing agents must be able to detect when novelty is injected, and to quickly adapt to the new rules.
We propose a model-based reinforcement learning approach where game state and rules are represented as knowledge graphs.
The knowledge graph representation of the state and rules allows novelty to be detected as changes in the knowledge graph, assists with the training of deep reinforcement learners, and enables imagination-based re-training where the agent uses the knowledge graph to perform look-ahead.
\end{abstract}

\section{Introduction}
\label{Introduction}
Artificial intelligence (AI) researchers have turned to computer games as test-beds for progress toward real-world AI environments~\cite{laird:aimagazine01}.
Games have complex state spaces, require real-time reaction and planning, involve frequent dynamic world changes, and adversarial entities. 
Reinforcement learning (RL) agents have been demonstrated playing board games, Atari games~\cite{mnih2015human}, first-person shooters~\cite{jaderberg2019human}, real-time strategy games~\cite{vinyals2019grandmaster} and multiplayer online battle arena games~\cite{berner2019dota} near (or above) human-level performance.
Despite their size and complexity, the games that researchers have targeted to date are only tractable with modern machine learning techniques because they are {\em closed-world environments}---the action space, and state space, and the transition dynamics of a world are permanently fixed for the life of the agent.
This is limiting as many real world environments experience moments of fundamental changes that must be adapted to. For example, in the stock market a new technology might make trades suddenly happen a lot faster, or for an autonomous vehicle in an emergency situation the rules of the road may change. 
If games are to be useful test environments for the real world we must look to games in which closed-world assumptions are violated.  

In this paper we consider the problem of detecting and adapting to {\em open-world novelty} in games. 
Open-world novelty is a change to the observations types, action types, environment dynamics, or any other change in the world mechanics that making it different from the world originally used to train the agent.
We refer to these changes as ``open-world'' novelty because these types of change to the world mechanics were never observed prior to the introduction of novelty. 
Open-world novelty differs from the more standard notion of novelty associated with the discovery of new observations during learning.

Open-world novelty commonly occurs in board games and card games, such as in the case of ``house rules'', where a particular group of people plays a common game in a slightly different way.
For example, certain cards may be removed from a deck or different rules may apply to players under a certain age. 
Consider, for example, the board game, {\em Monopoly}, which is famous for having numerous house rules.
House rules do not typically present too much of a problem for people who are familiar with the original rules and who must adapt to play under a modified rule set. 
For reinforcement learning agents, however, even a small rule change can result in catastrophically poor performance.

The problem of adapting reinforcement learning agents to open-world novelty is a uniquely challenging transfer task because agent policies are often fine-tuned to the mechanics of the environment.
Meeting this challenge can be decomposed into two primary tasks: detecting the moment and nature of the world mechanics change---what we refer to in this paper as the \textit{novelty injection}---and adapting prior models and policies to the world mechanics post-novelty injection. 

In this paper, we propose a deep reinforcement learning technique for detecting and adapting to open-world novelty in games by modeling the mechanics of the game world using knowledge graphs.
%
%
It is capable of learning the game rule, detecting open-world novelty by knowledge graph during the game play, and adapting to the novelty by imagination-based retraining with rule-based knowledge graph.
In our proposed technique, the knowledge graph encodes both state and how actions affect state changes.
Specifically, the knowledge graph plays three roles in our technique:
    (1)~it is used to quickly identify new rules;
    (2)~It can be embedded into the deep reinforcement learning network architecture to achieve faster policy convergence; and
    (3)~it can be executed by the agent to perform look-ahead at how the novelty injection affects the game.



Our method is compared to the baselines of a vanilla A2C agent and a static strong heuristic-based agent in multiple types of novelty situations. We will examine whether our technique can detect the novelty explicitly, yield superior post-novelty performance, and achieves this performance in a fraction of the required updates. 

\section{Related Work}
\label{Related Work}

Most work on game-playing reinforcement learning  considers novelty with regard to unseen observations during exploration.
One exception is Tomasev et al.~\cite{tomavsev2020assessing}, who trained the AlphaZero~\cite{silver2017mastering} system to play nine chess variants. 
However, they focus on automatically balancing game mechanics instead of detecting rule change and dynamically adapting models as in our work.

Model-based reinforcement learning~\cite{moerland2020modelbased, sutton2018reinforcement} techniques attempt to learn a model of state transitions---also called the {\em forward function}.
Techniques in games include: modeling the forward function with GANs~\cite{kim2020learning}, recurrent neural networks \cite{ha2018recurrent,tacchetti2018relational,freeman2019learning}, or inductively learned symbolic transition model~\cite{guzdial2017game}.
We adapt the inductive search for a forward model by Guzdial et al.~\cite{guzdial2017game} to generate a knowledge graph that relates the current state to successor states.
This technique, originally developed for modeling video games, learns to identify the features that are necessary before an action can be performed (pre-conditions) and the features of the environment that change after an action is performed (post-conditions) via repeated observations.
This technique was demonstrated with value iteration techniques to solving Markov Decision Processes but not reinforcement learning.

The way we use the knowledge graph to represent the state shares some conceptual similarity to Tacchetti et al.~\cite{tacchetti2018relational}, who propose Relational Forward Models (RFM) in which they use relational graphs to model world world state.
RFMs encodes the world state as a graph and then learns an RNN to predict what the state graph will look like in the future. 
They apply their technique to playing StarCraft II~\cite{zambaldi2018relational}. 
Our work is distinct in that we represent entity relationships in the knowledge graph as well as the state, and are using rule-based models for the forward function, which suffer from fewer learning challenges (like vanishing and exploding gradients) and are easier to train, not requiring the many updates to maintain both an updated state graph and graph-predictive RNN.

The problem of adapting to open-world novelty is related to {\em transfer learning}.
Transfer learning is characterized by a problem wherein an agent must perform well in task $T_n$ by transferring some knowledge from tasks $T_1, T_2, ... T_{n-1}$.
Few-shot transfer learning is the problem of reducing the number of training steps in $T_n$.
One can consider the novelty injection as a transfer from task $T_i$ to $T_{i+1}$. 
However, the agent is not told  the transition has happened, it must first identify the novelty in its environment and then adapt to it.

{\em Advantage Actor Critic} (A2C)~\cite{mnih2016asynchronous} agents solve Markov Decision Processes (MDP) by learning a policy through trial-and-error. 
For each step the agent takes in the environment observation, the agent uses a neural network with two heads.
The ``actor'' head takes in the observation $\mathbf{s}_{\mathbf{t}}$ and from its policy $\pi_\theta (a_t|s_t)$ computes the next best action $a_t$.
The ``critic'' head then takes in the next state vector $\mathbf{s}_{\mathbf{t + 1}}$ as computed from action $a_t$ and estimates the {\em value} of the action given the state. 
The policy gradient of A2C can be computed as,
$\nabla_{\theta} J(\theta)=\mathds{E}[\nabla_{\theta} \log \pi_{\theta}(a \mid s) A(s,a)]
\label{eq:a2c}$,
where $A(s,a)=r+\gamma Q^{\pi_{\theta}}\left(s^{\prime}\right)-Q^{\pi_{\theta}}(s)$ refers to  advantage value, $r$ is the reward, $\gamma$ is the discount factor, and $Q^{\pi_{\theta}}(s)$ represents state value given by ``critic'' head. 

KG-A2C~\cite{ammanabrolu2019graph} is a variation of the standard A2C reinforcement learning technique that uses an embedded knowledge graph of the environment to augment the state and constrain the agent's policy. A \textit{knowledge graph}, $KG = \{\langle subject, relation, object\rangle\} \subseteq E \times R \times E$, is a graph-structured knowledge bases, composed of entities (nodes) and relations (different types of edges), where $subject, object \in E$ are two entities, referred to as the subject and object, and $relation \in R$ is the relationship between two entities. Furthermore, the ``concepts'' that comprise the nodes in the knowledge graph can be drawn from a large vocabulary of natural language words or composed from sub-words. 
They were initially developed for  text-based games~\cite{ammanabrolu2019playing,ammanabrolu2019graph} which emphasize discovery of previously unobserved parts of the world state as the world is explored.
We build off the KG-A2C framework.

\section{Proposed Approach}
\label{approach}

Consider the problem where a reinforcement learning agent has trained on game modeled as a partially-observable Markov Decision Process (POMDP): $G=\langle S, P, A, O, \Omega,R,\gamma\rangle$, 
representing the set of environment states, conditional transition probabilities between states, actions, observations, observation conditional probabilities, reward function, and discount factor, respectively. 
The agent has learned a policy $\pi_G(o)\rightarrow a$ by playing $t$ complete games.
At time $t+1$ the agent is introduced to game $G'=\langle S', P', A', O', \Omega',R, \gamma\rangle$ where and of $S'$, $P'$, $A'$, $O'$, or $\Omega'$ may or may not be distributed equivalently  to their counterparts in $M$.
At the start of the $(t+1)^{\rm st}$ game, we say that novelty has been ``injected''.
Further suppose that the exact nature of the novelty is unknown and that the agent is not explicitly told whether it is playing $G$ or if one of the nearly infinite number of possible $G'$ games.
The agent must (1)~detect the novelty, and (2)~update $\pi_G(o)$ to $\pi_{G'}(o)$ as quickly as possible. 

\begin{figure}[t]
    \centering
    \includegraphics[width=1\linewidth]{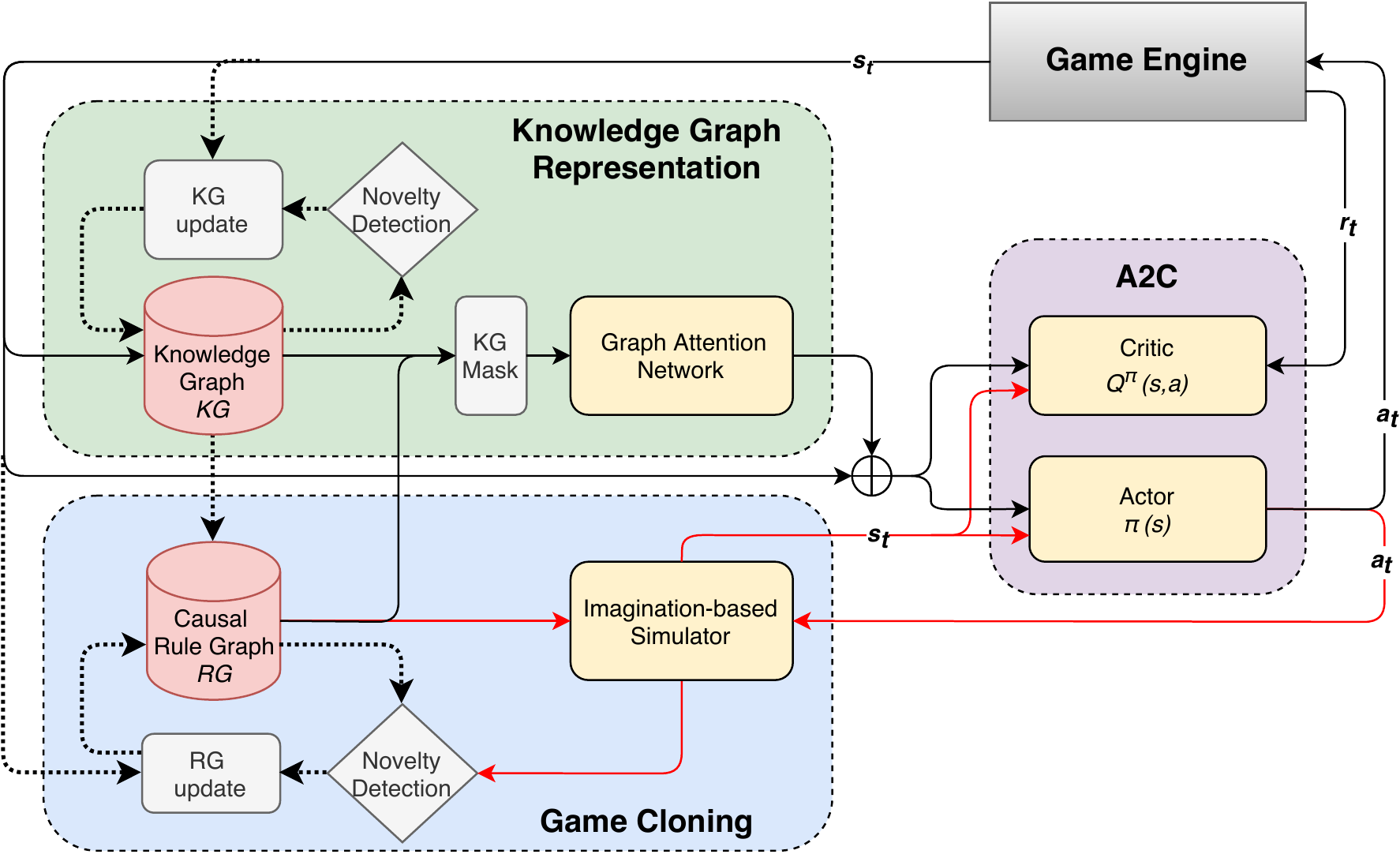}
\caption{The proposed architecture. The agent receives observations from the external environment (black pathways) or it's internal rule knowledge (red pathways) depending on whether it is pre-training, testing or re-training after novelty injection. }
    \label{fig:architecture}
\end{figure}

Figure~\ref{fig:architecture} shows the architecture of our approach to solve the open-world novelty problem.
Our system maintains two knowledge graphs: one that represents the state of the world (KG) and one that represents the rules of the game (RG). 
The KG is further divided into two parts: dynamic state, which is comprised of entities and relations that can change after any given step in the game, and static state, which is comprised on entities and relations that do not change.
The architecture provides for several pathways.
After every step the agent continues to learn the static portions of the state KG and the rule graph (dashed lines); these pathways are always active.
The black solid lines represent the standard loop used by the agent when playing the original game $G$.
However, when novelty is detected, the agent switches first to the dashed line pathways to update the rule and knowledge graphs, and then to the red pathways to update its policy model based on its internal model of the new rules.
Notice that it does not interact with the game engine in this mode, instead using it's rule graph as a surrogate for $G'$, training itself on its {\em imagination} of the new game.

\subsection{Novelty Detection by Knowledge Graph}
\label{sec:nd}

Our proposed technique learns the rules of the game, which describes how the dynamic portions of the game state change when actions are taken, as well as the static elements of the game state. 
For example in the game of {\em Monopoly} a static element of the state is the price of Baltic Avenue is $\$80$ and a rule is that rolling a 5 from Baltic Avenue results in the player's location changing to Vermont Avenue. 

We assume that the agent has played a number of games to discover a complete set of rules and static state entities and relations.
The agent continues to monitor the rules and static entities after every step in the game.
After every state change in the game, we extract entities and relations from the observation.
Our system detects novelty as a change to static entities and relations of the state knowledge graph or to the rules graph.
Changes include a new node becoming linked to part of the original graph, an original node  becoming un-linked, or a new relation linking existing entities.
Some rules involve distributions of values, such as die roles or proportions of cards in a deck.
Detection of these types of novelty---such as the number of sides of a die---requires monitoring the distribution of values until the agent has high confidence that the distribution has shifted. 

When novelty is detected, we expect to see the agent's performance on the game to drop significantly (though not always, sometimes the novelty can be irrelevant to the policy).
Novelty detection unfreezes the agent's policy to continue learning.
The agent is ``online'' in the middle of a game and must revise its policy in as few turns as possible;
the agent can train in between turns and also learn from its own online moves.

\subsection{Reinforcement Learning with Knowledge Graphs}
\label{sec:kg-a2c}

Our architecture maintains two separate knowledge graphs, the world state knowledge graph (including dynamic and static components) and the rules graph.
Our model embeds the world state knowledge graph and game observation state and passes the combined embedding to the ``actor'' and ``critic'' heads of an A2C network. 

Firstly, world state knowledge graph is embedded and passed into Graph Attention networks (GAT). Details are as follows. The input features, $h = \{	\overrightarrow{h_1},	\overrightarrow{h_2},...,	\overrightarrow{h_N}\}, 	\overrightarrow{hi} \in R^F$
are transformed into higher-level features by a linear layer, $W \in R^{{F'} \times F}$, where $N$ is the
number of nodes, and $F$ is the number of features in each node. Then a shared attention mechanism,  \textit{self-attention}, is applied to all the nodes. It first computes attention coefficients, indicating the importance of node $j$’s features to node $i$.
\begin{equation}
e_{ij} = LeakyReLU(\textbf{p} \cdot W(h_i \oplus h_j))
\end{equation}
where $\textbf{p}$ is a learnable parameter. Then attention coefficients $e_{ij}$ are normalized to $\alpha_{ij}$ because each node can have a different number of neighbours. 
\begin{equation}
\alpha_{ij} = \frac{exp(e_{ij} )}{\sum_{k\in \mathcal{N}_i}{exp(e_{ik})}}
\end{equation}
where $\mathcal{N}_i$ is neighborhood of node $i$ in the graph. 
Then the final world knowledge graph is embedded into $\mathbf{g}_{\mathbf{t}}$.
\begin{equation}
\mathbf{g}_{\mathbf{t}}=f\left(W_{g}\left(\bigoplus_{k=1}^{K} \sigma\left(\sum_{j \in \mathcal{N}} \alpha_{i j}^{(k)} \mathbf{W}^{(k)} \mathbf{h}_{j}\right)\right)+b_{g}\right)
\end{equation}
where $k$ refers to the parameters of the $k$th independent attention mechanism, $W_g$ and $b_g$ are the weights and biases of the output linear layer, and $\bigoplus$ represents concatenation. 

Secondly, the game observation state is transformed to $\mathbf{o}_{\mathbf{t}}$ by a linear layer with parameters, $W_o$ and $b_o$. 
Finally, the final observation vector $\mathbf{o}_{\mathbf{t}}$ and the final world knowledge graph embedding vector $\mathbf{g}_{\mathbf{t}}$ are concatenate to be the input of A2C model, $\mathbf{s}_{\mathbf{t}} = \mathbf{g}_{\mathbf{t}} \bigoplus \mathbf{o}_{\mathbf{t}}$.

Encoding rules in the knowledge graph helps in two ways.
First, it provides context to the current state and allows the agent to attend to the context surrounding the state.
There may be concepts related to the current state that are not explicitly part of the current state but provide additional patterns that inform decision making.
Second, concepts that the agent may encounter one step in the future are linked to the current state, providing the agent with one-step look-ahead when deciding on the action to perform (similar to relational forward models~\cite{tacchetti2018relational} but looking at all future states instead of one most likely state).

We feed only a portion of the rules graph and static state knowledge graph into KG-A2C for embedding.
We identify concepts relevant to the current state and filter out all nodes in the graph farther than $k$ edges.
If the entire rule graph is fed in, then the graph never changes after all the rules are learned and KG-A2C learns to ignore the graph.
In essence, we are augmenting the current, observable state with non-observable related concepts, some of which are relevant to the agent's immediate future.

We will introduce the preliminary experimental results of our knowledge-graph based reinforcement learning in the Monopoly game domain with novelty injections in following Experiments Section.

\subsection{Game Cloning and Imagination-Based Retraining}
\label{sec:cloning}
In this section we describe our rule learning technique, which is shown as then blue box in Figure~\ref{fig:architecture}.
Modeling the forward function is well suited to the problem of novelty adaptation because a small change in the environment may have a massive effect on the agent's policy. As prior work has shown in other non-novelty related domains, a forward model can be learned from data as in the case of AlphaGo and AlphaZero \cite{silver2016mastering}. However, neural networks as forward models can also take a large amount of data to learn. By making the forward model a symbolic rule-based model and combining it with the knowledge graph's ability to detect novelty in a precise, grounded way, we can update a symbolic model much faster. Then, with an updated forward model, even while playing a game, the agent can play through entire ``imaginary'' games using the forward model, and then update its model independently of the game it is currently playing.

The rule graph is a bipartite directed causal graph that represents the full set of rules that govern a state transformation.
Game rules are IF-THEN causal relationships between different relational 3-tuples from the world state knowledge graph, which are in the form of $\langle subject, relation, object\rangle$.
Each rule consists of a set of preconditions, which are relational tuples that must be true for the rule to take effect, an action, which functionally is a ``special'' precondition that is not a tuple, and a post-condition change, which is a change in a KG relational tuple that results if all of the preconditions are met. 

The core purpose of the rule graph is to predict the next state given the current state and action.
Because the rules graph tells us how the state KG changes, the agent can simulate the game instead of using the real game engine.
For the purposes of learning, however, we collect a dataset of state, action, next state tuples $(s,a,s')$ from game play. 
The prediction function returns the potential next frames, accounting for the stochasticity of the game and different player actions, each with a likelihood. 
The state with the highest likelihood is the predicted next state.
If the prediction accuracy is less than a certain threshold for a pair of states, we run our rule graph update algorithm on the current rule graph, $(s,a,s')$ tuple, and KG. 
The graph update process is shown in Algorithm~\ref{alg:rule-learning}, which is adapted from the game engine search algorithm of Guzdial et al.~\cite{guzdial2017game}, originally designed to find a set of rules for how sprites evolve in 2D graphical games. 
The update involves searching through \textit{neighbor rule graphs}, or graphs that are the same as the current graph except for a small change.

Our algorithm considers three types of neighbor rule graphs:
(1)~Adding a novel KG tuple as preconditions to an existing rule,
(2)~relaxing an existing rules' set of preconditions given the intersection of the current knowledge graph relations and current preconditions,
(3)~adding a new rule to the rule graph. 
Novel knowledge graph tuples occur when the knowledge graph has entities or relations added that have not yet been present in the knowledge graph. These are simply added appropriately to the preconditions and post conditions of the rules in the neighboring rule graphs. Relaxing the existing rules' preconditions is similar, but instead of adding neighboring rule graphs include rules with preconditions removed, effectively making the rule less strict. Lastly, adding rules to the graph requires picking out a pair of tuples of the same relation and that concerns the current state and next state, but differ in their subject or object. The rule is then initialized with all of the current state-related knowledge graph tuples as preconditions and the change in the tuple becomes the postcondition change.

\begin{algorithm}[t] 
\SetAlgoLined
\KwData{$D(), \epsilon>0, R, G, s, a, s' $}
\KwResult{$R^{*}$}
$C \leftarrow []$\;
$P \leftarrow P \cup [R,\infty]$\; 
$d* = \infty$\;
\While{$P \ne \emptyset$}{
    $\hat{R} \sim P$\;
    $d = D(\hat{R}, s, a, s')$\;
    \eIf{$d < \epsilon$ }{
        return $R^{*}$\;}{
        $N = C \cap \mathbf{NeighborGraphs}(\hat{R}, G)$\;
        $P \leftarrow P \cup (N \oplus [\cdot,d])$\;
        \eIf{$d* > d$}{
            $C \leftarrow C \cup R$
            $R \leftarrow \hat{R}$\;
            $d* \leftarrow d$\;
            }{
            $C \leftarrow C \cup \hat{R}$\;}
        }
    }
$R^{*} = \hat{R}$\;
\caption{Rule Graph Update}
\label{alg:rule-learning}
\end{algorithm}

In Algorithm \ref{alg:rule-learning}, in addition to the $(s,a,s')$ tuples for prediction, we require as input:
\begin{itemize}
\item $D$, the distance function between prediction and ground truth next state, which runs the $Predict$ function described above,
\item $R$, the current rule graph ,
\item $G$, the current knowledge graph, and 
\item $\epsilon$, the convergence threshold. 
\end{itemize}
In this algorithm, $C$ is the set of examined rule graphs neighbors, $P$ is a priority queue representing the set of unexamined neighbor rule graphs, and $d*$ is the closest prediction distance. While there are unexamined neighbor rule graphs and prediction distance is below the threshold, we examine neighbor graphs $\hat{R}$ according to $P$. If they predict the next state with a prediction distance below the threshold the new graph is immediately returned. If it is not below the threshold, but it is closer than the current graph, the neighbor graph becomes the current graph and the previous current graph is moved to the closed set. If not, the neighbor is moved to closed set and the next neighbor is selected in the next iteration of the loop. The neighboring engines in $P$ are sorted by the distance to their parent so that there is no need to simulate the next frame and calculate the distance until an opened neighbor graph is observed. This continues until a rule graph that can predict the next state with an accuracy below the threshold or the priority queue of neighbors is empty, in which case the best rule graph is returned. The rule graph update algorithm is, therefore, essentially a priority-based search over neighboring rule graphs. 

After novelty is encountered and the graph is updated adapted, we then can use this learned rule graph to simulate gameplay for updating the reinforcement learner with {\em imagination-based simulation}. The agent continues to play the game online, interacting with the game engine. 
In between each move the agent uses the rule graph to run as many ``imaginary'' trials offline as it has time for. Not only can we use this for updating agents online, but we can focus on agent performance in states that are, according to the knowledge and rule graphs, most relevant to the recently adapted novelties. This results in faster online re-convergence of the policy model because the agent can run many more trials in its internal model of the game engine than would be feasible online.

\section{Experiments}
We illustrate our open-world novelty agent running within the implementation of \textit{Monopoly}, which is a game where players compete to earn money and bankrupt opponents by buying, renting, and selling properties. 
Monopoly is a game that well known for having a lot of ``house rules'', and is a good experimental test-bed for open-world novelty because it is relatively easy to modify to create novelty.
One can change the color of properties, the number of properties, how to get out of jail, among other things, and the game will still work but an agent with a policy optimized for the original rules may fail using the new rules. 
As a concrete example, we can imagine if the purchase price of 50\% of the properties might increase to 5 - 10 times higher than original prices, an agent that learned a policy with the original property rules would quickly bankrupt itself in a game using the new property rules, and so it must adapt its policy to accommodate this rule change. 
Since Monopoly is a multi-player game, we play against a rule-based heuristic agent that is engineered to play a  strong game according to conventional wisdom. 

We use this Monopoly environment to conduct experiments to (1) evaluate the novelty detection technique, (2) observe whether the knowledge graph-constrained reinforcement learner adapts to novelty with fewer retraining iterations than standard reinforcement learners, and (3) evaluate the predictive accuracy of the rule graph-based game cloning.

\subsection{Novelty Detection Evaluation}
\label{sec:exp_nd}
To evaluate our novelty detection algorithm, an independent research team at the University of Southern California Information Science Institute, created $n$ versions of Monopoly, each with a different novelty. We were not provided any information on what novelties were invented other than a very broad description, i.e. any rules or attributes about board might be changed.
Our system successfully detected $73.8\%$ of the novelties and only took average $5.86$ steps to detect each novelty.
When novelties were detected, it was done in a single step; the average of $5.86$ factors in the length of games where novelties were not detected.
Our system produced no false positives where novelties were detected when none were injected.
Because of ongoing single-blind novelty detection evaluation, the novelties that we failed to detect remain secret so that we cannot specialize our novelty detection technique and artificially inflate our results at a later time.
However, we note that our win-rates on games where we failed to detect novelty did not always drop, indicating that the novelties we did not detect may not required the agent to change it's policy.

\subsection{Reinforcement Learning with Knowledge Graphs Evaluation}

In this experiment, we only consider the parts of our system architecture involved in the state knowledge graph (the green and purple boxes in Figure~\ref{fig:architecture}), referred as \textit{KG-A2C}. 
As a baseline, we compare our system against a vanilla A2C agent.
Vanilla A2C only considers the Game Engine state as input and only has the actor and critic as learnable components (indicated by the purple box in Figure~\ref{fig:architecture}). KG-A2C on the other hand utilizes a knowledge graph representation and a learnable Graph Attention Network to augment the state (indicated by the green box in Figure~\ref{fig:architecture}.


For this evaluation we pre-train KG-A2C and vanilla A2C agents on Monopoly playing against the rule-based heuristic agent. We only consider the offline training here to demonstrate that incorporating knowledge graph into A2C is able to help the agent more quickly adapt to the new rules. 
At the start of the evaluation, we inject novelty by randomly setting the purchase price of $x\%$ of the properties to \$1499 or the rent of $x\%$ of the properties to \$0. 
We then keep retraining the baseline and KG-A2C agents. We measure the win rate before and after novelty injection when playing against the rule-based agent. We test the agent on the same 1000 games after every 300 steps.

\subsubsection{Results}

\begin{figure}[H]
    \centering
    \subfloat[50\% property price raised to \$1499]{
	\label{fig:sfig1}
	\includegraphics[width=.75\linewidth]{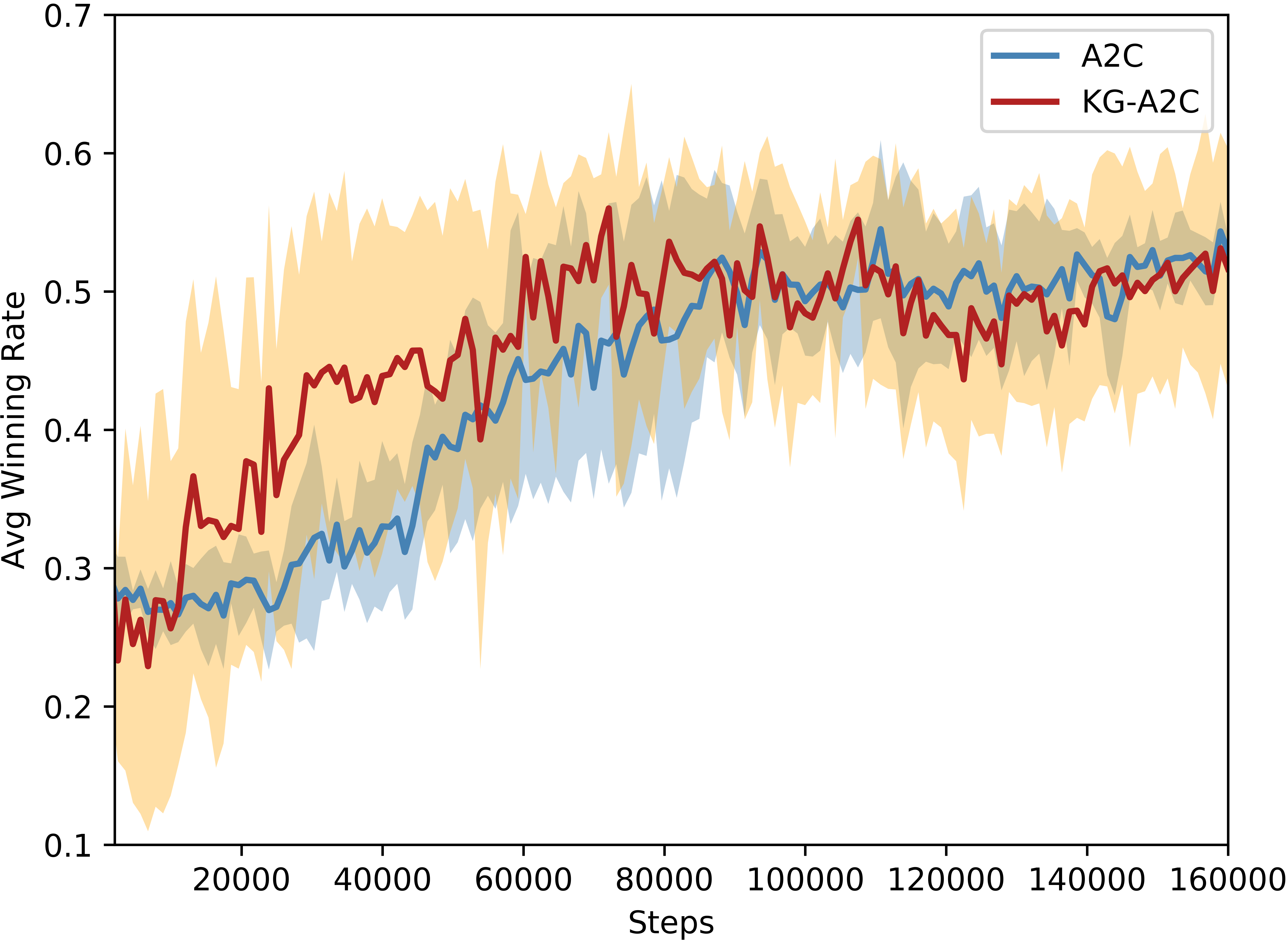}}\\
	\subfloat[70\% property price raised to \$1499]{
	\label{fig:sfig2}
	\includegraphics[width=.75\linewidth]{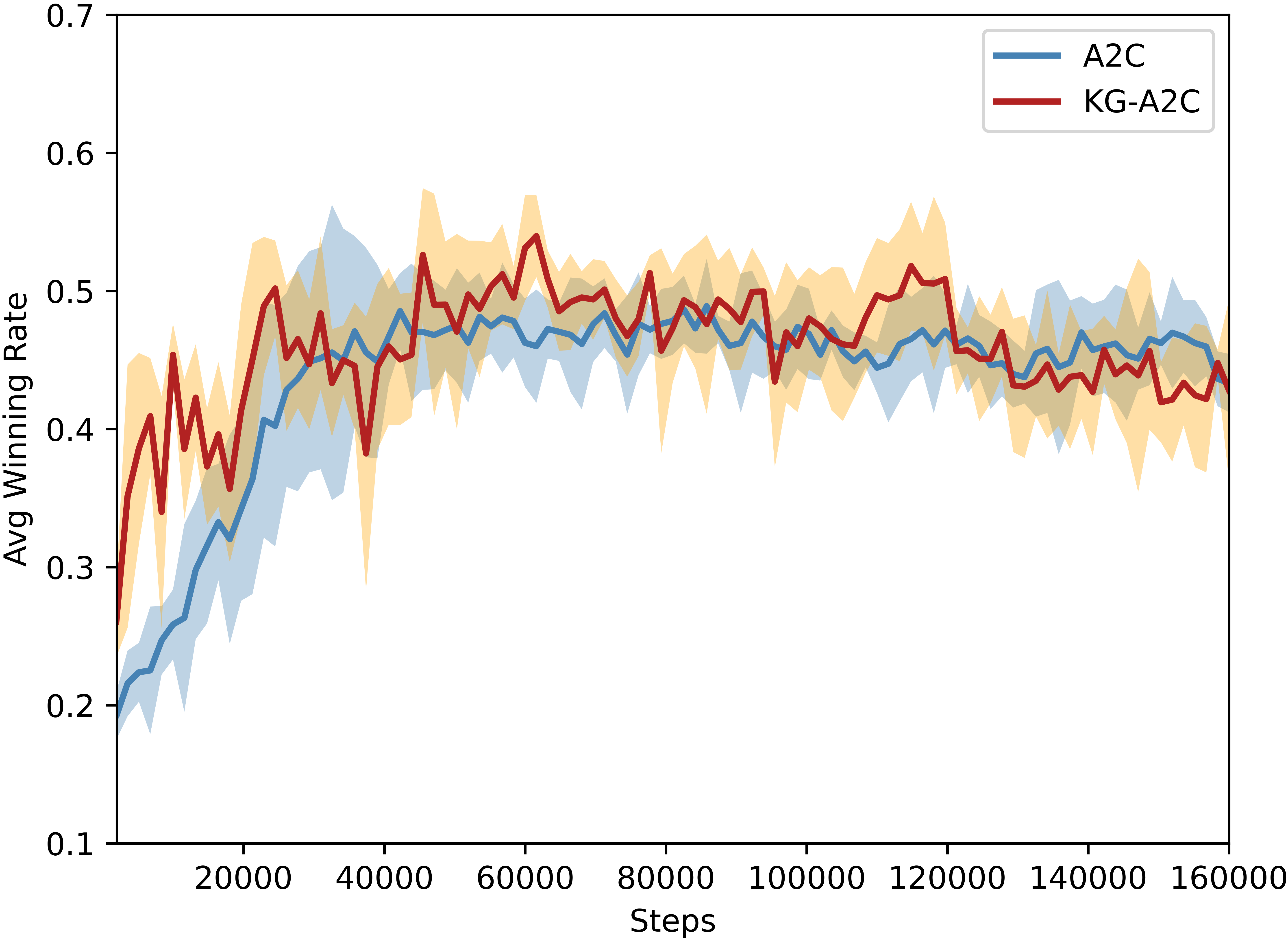}}\\
	\subfloat[50\% property rent decreased to to \$0]{
	\label{fig:sfig3}
	\includegraphics[width=.75\linewidth]{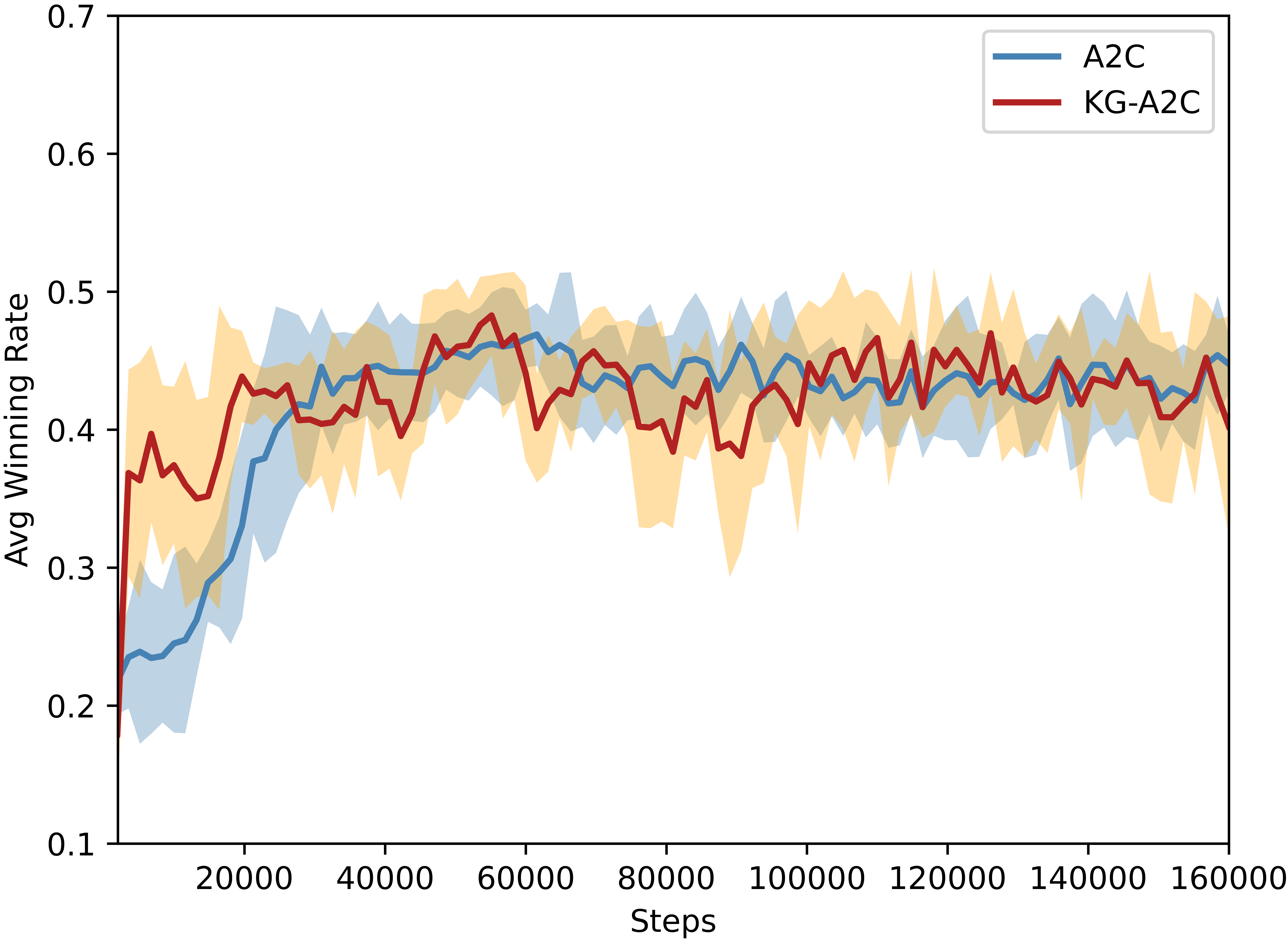}}
	
    \caption{Average winning rate of offline training tests after novelty injection, averaged across 5 independent runs. Orange and light blue show the confidence intervals for KG-A2C and A2C, respectively.}
    \label{fig:kga2c_exp}
\end{figure}

The results are shown in Figure~\ref{fig:kga2c_exp}, where the solid line represents the mean of \textit{window win rate} over 5 independent runs and the filled light color represents the confidence interval\footnote{We take $\mu - \sigma \leq X \leq \mu + \sigma$ as confidence interval.}.   

Prior to novelty injection, both the baseline A2C agent and our novelty-aware KG-A2C agent have a win rate of $\sim$48\% against the rule-based agent. 
The winner of Monopoly between two competent players is primarily determined by lucky die rolls and learning-based agents do slightly worse than an agent with hard-coded with heuristic knowledge about the game.
When novelty is injected the win rates against the rule-based agent drop significantly, to $\sim$25\%. 
Our novelty-ware KG-A2C agent and the baseline A2C agent both re-converges to a $\sim$50\% win-rate against the rule-based agent.
The significant difference between novelty-aware KG-A2C and  the  baseline  A2C  is  that  our  agent  re-converges  significantly  faster.  The results show offline retraining performance, although online performance where the agent must play the current game while also training will present a similar result. Were this online learning, the novelty-based KG-A2C is likely to provide better moves at any time while it is re-converging; it’s average performance post-novelty is higher than baseline, but also it’s lower confidence bound is often no worse than that of the baseline.

\subsection{Game Cloning Evaluation}

To evaluate the game cloning component of our work we extract state samples from Monopoly gameplay and measure the accuracy of our rule graph's prediction of the next state. Specifically, we save the full game state as a dictionary--the same format that is the input to our KG-A2C--at the start of each players turn. This is slightly different than the input frequency of the A2C models, which observe the game state only at the start of the agent's own turn. While the actor and critic functions can abstract away the other players and board actions as part of the environment, the dynamics of the players and the board are more easily modeled when learning a rule graph that can be applied to each player separately. Therefore we simply reduce the step of the rule graph-base model for learning, and when using the Imagination-based Simulator for hypothetical simulation it takes one step for each player, and aggregates the changes to predict the next state for the KG-A2C model.  

The gameplay data is separated into train and validation sets. The training set consists of $200$ consecutive turns of a single continuous Monopoly game. This data is observed in order as that is representative of how the rule graph will encounter data as part of the KG-A2C training process. The validation data is also extracted from a game, but then split into tuples of state, action, next state, $(s, a, s^{'})$, and then shuffled. This is because we are using this data to evaluate the one-step predictive accuracy of the rule learner at each step, so the order of the validation states is unimportant.  
 
We evaluate the accuracy of next-state prediction using a heterogeneous \textit{prediction distance} that allows us to account for the continuous and discrete values in the state, where all values are scaled down so that an ``adjacent'' change is equal to 1. Binary values are 0 for the game-start state and 1 otherwise, for example whether a property is mortgaged ($1$) or not ($0$), and along with finite unordered set values use an ``overlap'' distance, the indicator function, where if the values are equal the difference is 0, otherwise it is 1. 
This applies, for example, when a property changes from one owner to another. 
Finite ordered values are measured by the absolute difference between the values; for example if a house is added to a property, this changes the amount of infrastructure on that property by $+1$. Continuous finite values are made discrete ahead of time by binning so that the smallest single action yields a difference of 1. For a infinite values, we measure the Euclidean distance metric divided by the original value, which is in effect the percent Euclidean change. If the value type is not covered by these cases or is unknown, a difference of 1 is always returned. The final prediction distance is then the square root of the sum of the squares of each of these value differences, or

\begin{equation}
    D(s,\hat{s}) = \sqrt{\sum_{i=1}^{m} d_i(s_i,\hat{s}_i)^{2}}
\end{equation}

where for each $i \exists S_i \text{ s.t. } s_i \in S_i,$
\begin{equation}
d_i(s_i, \hat{s}_i) = \begin{cases} 
      \mathds{1}[s_i, \hat{s}_i] & S_i \text{ unordered} \\
      \frac{\|s_i+\hat{s}_i\|_2}{s_i} &   |S_i|=\infty \\
      |s_i - \hat{s}_i| & 2<|S_i|<\infty \\
      1 & \text{otherwise}
      \end{cases}
\end{equation}
which defines the difference between the state attribute values. This metric is the same that we use as the distance metric in Algorithm \ref{alg:rule-learning} to measure the quality of neighbor graphs. This is similar to the the classic Heterogeneous Euclidean-Overlap Metric (HEOM)~\cite{wilson1997improved}.

The rule graph update algorithm (Algorithm \ref{alg:rule-learning}) is an algorithm designed to run online, continually running and evaluating on the game in order. 
After the rule graph update from each training sample, we evaluate on the validation samples. 
The validation error per step of training is displayed in Figure \ref{fig:gamecloning}. For our evaluations we tuned the matching threshold $\epsilon$ for performance, and the chosen value was $\epsilon = 4$. This value, however is sensitive to the problem set specifications, including the state distribution and the choice of distance metric.  We can see that after a few steps the average value quickly goes down, but remains above zero. We believe this is the result of not only the distance function and threshold but also the type of data we store in our knowledge graph. It indicates that our KG is either too sparse and missing information, or that it is so overspecified that the game engine is still overfitting to certain relations. In future work, we plan to analyze this unusual behavior and remedy it by further tuning of the threshold and distance functions, as well as a better specified knowledge graph.

\subsubsection{Results}

\begin{figure}[t]
    \centering
    \includegraphics[width=\linewidth]{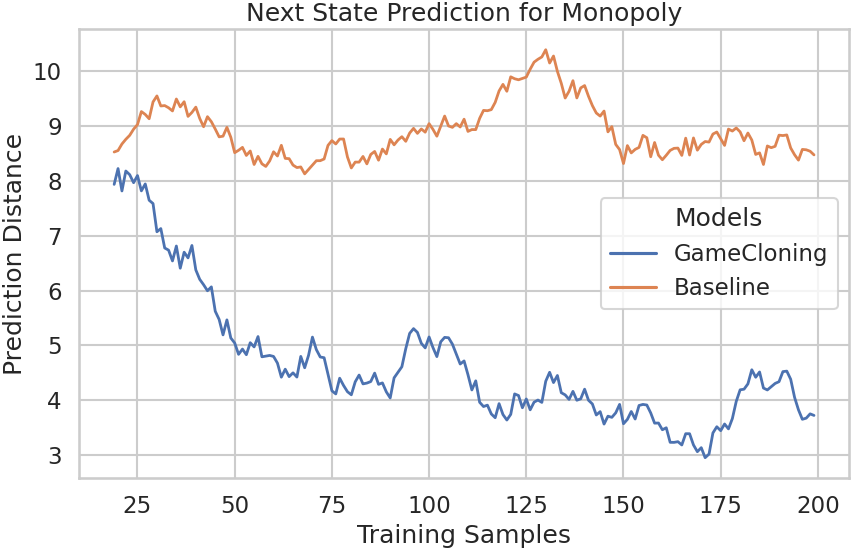}
\caption{Distance of next state prediction using the game cloning rule graph. The plot shows the prediction distance (error) of two prediction models, rule graph-based \emph{GameCloning} and the 1-step change random prediction baseline, \emph{Baseline} after a certain number of training samples. Both are plotted as their 20-sample moving averages. The prediction distance at each training sample is the average prediction distance on a held out validation set after the rule graph update algorithm terminates for that training sample.}
    \label{fig:gamecloning}
\end{figure}

The plot in Figure \ref{fig:gamecloning} shows our method and a random guess baseline with average validation Mean Squared Error against number of training samples. In this particular case we observed that the rule graph updated at every step. 
We can see that while the rule graph never has a consistent perfect evaluation score it does a much better than random job at predicting the state, and that it gets to that point with very few training samples. This is the power of this method: the fact that with very limited training samples it can learn a forward prediction model that generalizes to a held out validation game data in very few training samples. This data efficiency makes the method especially well suited to adapting to novelties injected into the environment.  This is the first test necessary before using the rule graph for imagination-based retraining of the KG-A2C model.

\section{Conclusions}
\label{Conclusion}

In this paper, we consider the problem of open-world novelty, where the rules of the environment---how the world changes---can suddenly change. 
We consider the problem through the lens of games, where a player may be faced with adapting their game-playing strategy to ``house rules'' that are slightly different from those they are experienced with.
These types of house rules generally do not present human players with significant challenges, but can cause reinforcement learning agents to fail catastrophically.
We specifically approach open-world novelty using model-based reinforcement learning where we model the rules of the game as a symbolic knowledge structure. 
While games help us take early steps toward designing new novelty-aware agents, open-world novelty is not specific to games; it occurs in any domain where a task is assumed to be in a closed-world environment but is situated within a larger open world. 


Our preliminary results show that this method will adapt more quickly to rule changes in games than a vanilla A2C learner, because of the knowledge graph's ability to model both the unobserved relationships in the state as well as the rules of the game. 

This problem presents many challenges and future research questions. Specifically, can an agent adapt to the new rules of a game, and then go back to the original rules with no retraining? Does the use of a knowledge increase the explainability of the agent, and therefore improve the ability of reinforcement learning agents to interact with humans? Does this model of using a knowledge graph to support a reinforcement learner help when information extraction and entity recognition cannot be mined directly from text? These are just some of the questions that this proposed domain and method open to the community. 

\bibliography{main.bib}

\end{document}